# Variational Approach for Intensity Domain Multi-exposure Image Fusion

Harbinder Singh[1], Dinesh Arora[1], Vinay Kumar[2]

[1]*Electronics and Communication Engineering Department, Chandigarh Engineering College, Landran, Mohali, India, harbinder.ece@gmail.com, dinesharora.ece@cgc.edu.in*

[2]*Electronics and Communication Engineering Department, Thapar Institute of Engineering & Technology, University, Patiala, India, vinay.kumar@thapar.edu*

**ABSTRACT**

Recent innovations shows that blending of details captured by single Low Dynamic Range (LDR) sensor overcomes the limitations of standard digital cameras to capture details from high dynamic range scene. We present a method to produce well-exposed fused image that can be displayed directly on conventional display devices. The ambition is to preserve details in poorly illuminated and brightly illuminated regions. Proposed approach does not require true radiance reconstruction and tone manipulation steps. The aforesaid objective is achieved by taking into account local information measure that select well-exposed regions across input exposures. In addition, Contrast Limited Adaptive Histogram equalization (CLAHE) is introduced to improve uniformity of input multi-exposure image prior to fusion.

1. **INTRODUCTION**

Digital camera having electronic imaging array and analog camera with film use nonlinear mapping that determines how radiance in the scene becomes pixel values in the image. In the early days of photography, photo-sensitive film was used to capture luminance variations present in the scene. The experimental setup for the formation of the latent image was carried out principally in the Kodak Research Laboratories at Rochester and Harrow in 1947. An overview of the reactions of the photochemistry of silver halide is presented in [1] and [2].

HDR image encodings come into play when full range of color and luminance values need to be captured. The HDR photography produces a single photo that is correctly exposed in both the dark and light areas. Quadratic objective function based on least square error was used by Debevec and Malik [3] to recover smooth and monotonic response function. Furthermore, to ignore saturated pixel values, the weighting function is used to give higher weight to exposures in which the pixel's value is closer to the middle of the response function. Single-precision floating point values are used to encode recovered radiance map. The approach proposed by Mitsunaga and Nayar [4] utilizes polynomial approximation for their response function. This technique is suitable for low-cost consumer equipments.

In the literature, several tone-mapping methods for converting real-world luminances to display luminances have been developed and fulfilling the fast growing demand of displaying HDR data on conventional display devices. Most tone-reproduction algorithms make use of photoreceptor adaptation [5,6] and simulation of real-world light conditions [7] to achieve visually plausible results. Local light adaption property of HVS is adopted in the local operators to correspond to the visual impression an observer had when watching the original scene, while the global operator are spatially invariant and are less effective than the local

Beside software solution, to increase the DR of real-time digital still imagery and video camera, a lot of computational capability and processing speed is required which is possible with the assistance of graphics processing hardware [4,8,9]. Although this is largely achievable, the hardware design approaches did experience some key issues associated with the development of the real-time HDR reconstruction: Theses are: computation time, computation cost, registration, sensor design and realization in hardware. Various solutions for designing the HDR image sensors have been proposed throughout the years [10, 11, 12, 13]. An overview of various HDR sensor design techniques is presented in the tutorial by El Gamal and Yang [14].

The exposure fusion is the alternative solution to HDR that fuse multi-exposure images into single image. Over the years, various fast and effective weighted average based exposure fusion approaches have been proposed. Among these guided filtering [15] based two-scale decomposition fusion approach [16], global optimization using Generalized Random Walks (GRW) [17] for fusion [18], and median filter and recursive filtering [19] based fusion approach [20], are producing fusion results with better quality. A more modern approach for dynamic environment was proposed by Li and Kang [20], in which histogram equalization [21] and motion estimation using median filter [22] are included.

Mertens et al. [23] has proposed fusion technique using multi-resolution approach [24] without extending the DR and tone-mapping of the fused image. This method used multi-resolution technique to blends input images based on quality measures e.g. saturation and contrast. Moreover, flash image can be included in the exposures to enhance the details in the fused image [23]. The performance of this multi-scale technique is dependent on the number of decomposition levels, i.e. the pyramid height. For better performance, larger images would have to be processed in a higher number of pyramid levels than smaller images. An automated exposure fusion approach using optimization of the pyramid height is the one of the recently proposed solution by Kartalov et al. [25].

As compared to others existing exposure fusion approaches, Kotwal and Chaudhuri [26] proposed a new fusion alternative in which optimization technique is attempted to estimate the best possible matte, which act as weight for the fusion purpose. In this approach, multi-objective cost function is developed, which provides an iterative solution using the variational method. The mates are adaptively derived from the data and the corresponding fused image at every iteration. Another matte-less solution was published by Raman and Chaudhuri [27], which used unconstrained optimization problem for the selection of locally high contrast pixels. In another approach, gradient information based visibility and consistency measures are utilized to merge static and dynamic scenes, and concluded that image gradients convey important information about the latent scene [28]. The work proposed in this paper experiments that the information in the resultant fused image can be controlled with the help of local information measure.

2. **BACKGROUND**

This chapter will give the overview of previous work related to imaging technology, HDR imaging, HDR software, sensor design, tone-mapping, inverse tone-mapping, image fusion and exposure fusion. This chapter will discuss the history, capabilities, and future of existing solutions to handle wide range of illumination conditions present in the natural scenes. This chapter will first survey challenges in recording outdoor and indoor scene that includes a directly visible sun and shadows, respectively. It reviews HDR imaging technology that can encode the color gamut and DR of the original scene into true radiance maps. This chapter provides a brief review of state-of-the-art methods available for sensor design, and the trade-offs for HDR sensor design. Chapter also discusses the difference between the standard digital image representation and the HDR imaging technology. To date, various tone-mapping algorithms have been proposed which prior to display reduces the tonal range of HDR data to match the capabilities of display devices at a minimal side effect on quality. This chapter gives the overview of different global and local tone-mapping methods. It will also briefly review the background work about range expansion in the case of inverse tone-mapping algorithms. Then, this chapter presents brief discussion on exposure fusion approaches available in the literature and indicates their advantages over HDR imaging. At the end of this chapter, chapter provides a brief review of edge-preserving filters, Multi Scale Decomposition (MSD) and their utilization in computer graphics, and image processing applications.

According to them, silver halide crystals are photoconductors which are activated by the photons. The photosynthesis process is used to break down molecules into smaller units through the absorption of light and it involves electronic and ionic migration. In a conventional camera, the photosensitive-film is first exposed to true radiance to form a latent image. The film is then developed to change this latent image into visible image. Film scanner is used for digitization of the film that converts the projected light into electrical voltage based on photo-sensitive. The electrical signal is then digitized, and stored on digital storage medium. The detailed description of photochemical process involved in analog cameras is provided in [29] and [30]. Although, analog cameras have greater DR than their digital counterparts and development process is used to limit or enhance the information retrieved from the exposed emulsion. Dodge-and-burn technique is commonly used to extract maximum information during the development process which acts as a tone-mapping step in analog photography. Dry developing method was developed by Applied Science Fiction and marketed to extract the full DR from a negative [31].

HDR image encodings come into play when full range of color and luminance values need to be captured. The HDR photography produces a single photo that is correctly exposed in both the dark and light areas. The process of generating HDR image from input multiple exposure images [32,33] recover true radiance values across images taken with conventional digital camera. Debevec and Malik [3] and Mann and Picard [34] proposed a HDR imaging to record the entire range of the scene radiances from different exposures that were acquired with a standard camera. The CRF recovered from differently exposed images is used to create HDR image whose pixel values are equivalent to the true radiance value of a scene.

Beside software solution, to increase the DR of real-time digital still imagery and video camera, a lot of computational capability and processing speed is required which is possible with the assistance of graphics processing hardware [35,36,37]. Although this is largely achievable, the hardware design approaches did experience some key issues associated with the development of the real-time HDR reconstruction: Theses are: computation time, computation cost, registration, sensor design and realization in hardware. Various solutions for designing the HDR image sensors have been proposed throughout the years [38,39,40,41]. An overview of various HDR sensor design techniques is presented in the tutorial by El Gamal and Yang [14].

Conventionally used CRT and flat panel have limited DRs spanning a few orders of magnitude, which is much lower than the DR of real world scenes, often less than 100:1. Therefore tone mapping is required to display HDR images on conventional display devices and printers [42]. The tone mapping process must consider the property of Human visual System (HVS) to reproduce LDR images from HDR images. Therefore, the ultimate goal of tone-mapping operators is to maintain original appearance of the tone-mapped HDR image on existing display devices whose capabilities in terms of DR are insufficient. Therefore, appropriate metrics based on human visual perception need to be accounted to depict their original appearance. Thus, the investigation of response of photoreceptors distributed on the retina with a finite resolution is very worthwhile in the context of tone-mapping operators. The reader is requested to refer [43,44,45] for detailed descriptions.

Orgden et al. [46] has proposed pyramid solution for image fusion. The pyramid becomes a multi-resolution sketch pad to fill in the local spatial information at increasingly fine detail (as an artist does when painting). The Laplacian pyramid decomposes input image into different spatial band that are considered as band-passed images preserving local spatial information [47]. Another multi-resolution based fusion which employs gradient map of the input images to yield a fused image with true information [48]. This approach takes into account the horizontal and vertical gradient maps for producing fused gradient map for each orientation and resolution. This gradient fusion approach utilizes Discrete Wavelet Transform (DWT) and Quadrature Mirror Filters (QMFs) in the reconstruction process. This approach was implemented for input data provided by multi-sensory arrays. Another multi-sensor data fusion technique which utilizes Total Variation (TV) [48] is the one by Kumar and Das [49].

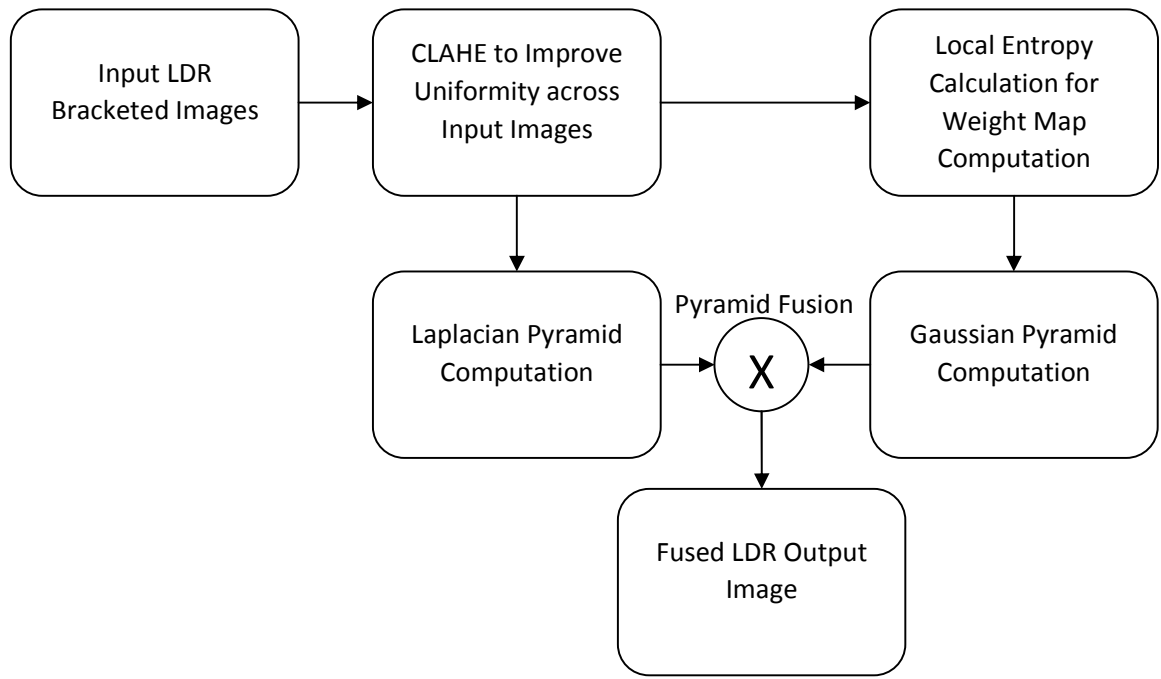

Fig. 1. Framework of proposed exposure fusion approach that utilizes CLAHE as preprocessing operator. The local entropy is computed from preprocessed images for weight function computation.

1. **PROBLEM FORMULATION**

The objective of present exposure fusion approach is to consider well exposed regions across input images taken with standard LDR cameras. It significantly improves information content in fused image for better human perception. The fused image can be directly displayed on LDR display devices without working in radiance domain [3]. In this paper, input multi-exposure images used are perfectly aligned to avoid unwanted ghosting artifacts due to objects and camera movements or captured from static scene with the help of tripod. Moreover, the final synthesized image does not require any post-processing.

As such, the present approach (See Figure 1) comprises following main steps:

1. CLAHE is introduced to handle non-uniformity across input multi-exposure image.

2. Local information measure is proposed for weight map computation that decides the contribution of pixels to the final fused image.

3. Multi-scale weighted average fusion is introduced for seamless blending.

## 2. METHODOLOGY

An intensity domain variational multi-exposure approach is proposed to overcome the limitation of standard LDR digital camera that does not record full range of light variations present in HDR scenes. It is found that pre-processing of input multi-exposure images based on CLAHE reduces inconsistencies introduced during image capturing. Even though such type of problem is addressed by Wu and Leou in [50], the Weighted Least Squares (WLS) optimization is applied on input LDR images for detail enhancement across input images. Thus, it is preferred to apply pre-processing operator prior to fusion so as to preserve details present in brightly illuminated and poorly illuminated regions of the HDR scene. In present approach, CLAHE is used for handling inconsistencies introduced by the nonlinear process of mapping of light intensities to pixel values when photograph -Linear Transformation for handling inconsisties.

Our algorithm is based on pixel level fusion approach which combines information from non-linearly transformed *N* different exposures. The process of arithmetic combination is used to blend corresponding pixels across input multi-exposure images. Arithmetic fusion can be summarized by the expression given as:

$$F(x, y) = W_1 I_1(x, y) + W_2 I_2(x, y) + C$$

where $I_1$, $I_2$, and *F* denotes input multi-exposure images and blended output image respectively at location *(x,y)*. $W_1$ and $W_2$ represent the weight function computed across individual input images that controls the contribution of input images in the fused image [51-54], and *C* the mean offset.

Computation of mean image across input images is the one example of arithmetic fusion technique. The output image is computed by taking the mean of input multi-exposure images; i.e., *$W_1$=1/2, $W_2$=1/2* and *C=0*. However, despite being significantly more computationally efficient than most other fusion systems, image averaging and other arithmetic fusion methods does not achieve enviable performance. The main reason for this is the loss of contrast, a result of destructive superposition when input signals are added. Further contrast reduction is also introduced when a normalized sum is used, such as in image averaging. In general, computation of mean image yields reasonable image quality in areas where pixel values are identical but the image quality rapidly decreases in regions where pixels values are completely different. The worst outputs are obtained in areas where the pixel values in input mages are photographic negatives of each other and contrast is annihilated.

In our algorithm, the corresponding pixels present in input images captured at variable exposures are fused in intensity domain. During fusion a non-linear intensity transformation based on CLAHE denoted by $T$ is utilized for dynamic range manipulation prior to fusion. The operator CLAHE is used as pre-processing tool. CLAHE utilizes adaptive approach for choosing local histogram mapping function. As compared to non-adaptive histogram equalization technique it yields better output image at extreme low and extreme high intensity variations. To compute local mapping function, the original input image is divided into various sub-images or local regions (i.e. 8×8 pixels). After dividing the input images into sub-images, histogram clipping is done using the local intensity mapping function. For computing adaptive mapping function local histogram is calculated for each sub-image instead of global histogram. After local histogram calculation and clipping it dynamically, bilinear interpolation is used to assemble neighboring tiles that overcome the problem of generation of artificial boundaries in processes image. It handles homogeneous regions (i.e. under-exposed and overexposed regions) across input multi-exposure images adaptively and yields optimal contrast in the output image. In proposed approach Rayleigh distribution is implemented as the intensity mapping function in CLAHE instead, which is defines as:

$$y_k = y_{\min} + \left[ 2\alpha^2 \ln \left\{ \frac{1}{1 - cdf(g)} \right\} \right]^{0.5}$$

where $y_k$ represents output intensity level for $kth$ ($1<k<N$) image, $y_{min}$ denotes low bound, $\alpha$ is a free parameter used for distribution and the value was set to 0.4, $g$ denotes the intensity level of input pixels, and $cdf(g)$ denotes the cumulative probability distribution computed across input image. Probability density function can be computed as:

$$P(y) = \frac{y_k - y_{\min}}{\alpha^2} \exp \left\{ -\left( \frac{y_k - y_{\min}}{2\alpha^2} \right) \right\} \quad for \ y_k \geq y_{\min}$$

Rayleigh distribution [55] is a continuous distribution, and sigma (σ) is used as scale parameter. It can be defined as:

$$P(x) = \frac{X}{\sigma^2 \exp(-X^2/2\sigma^2)}$$

The Rayleigh cumulative distribution function can be defined by following equation:

$$cdf(X) = 1 - \exp\left(-\frac{X^2}{2\sigma^2}\right)$$

where $\sigma$ is the user defined parameter which has positive values.

A. *Local Entropy*

Entropy is a statistical measure of randomness or uncertainty. In proposed approach it is utilized for weight map computation. It used to analyze the texture details present in the input images. It can be defines as:

$$H = -\sum_{G=0}^{255} P(G) \log_2(P(G))$$

In the above expression, *P(G)* denoted the probability of the occurrence of intensity value G, and the term *log2* is the base 2 logarithm that indicates the amount of uncertainty associated with the corresponding intensity value. This local measure can be considered as metric of information gained within the local window for the computation of weight function.

In our approach local entropy will be different from region to region across input images captured at variable exposure settings. Well exposed regions yields higher value of local entropy as compared to the highlights and shadows. To compute local entropy, the input image is divided into small blocks of window size 3x3. The local entropy for *k-th* input image within 3x3 local windows is defined as follows:

$$H_{i,j,k} = -\sum_{I'=0}^{255} P_{i,j,k}(G) \log_2(P_{i,j,k}(G))$$

$$W_{i,j,k} = \left[-\sum_{k'}^{N} H_{i,j,k'}\right]^{-1} H_{i,j,k'}$$

where *Hij,k* represents entropy value of the 3-by-3 neighborhood of each block in the *k-th* *(1<k<N)* input image (i.e. total number pixels having intensity divided by the total number of pixels within the local neighborhood) and *Wij,k* is the Normalized local entropy (i.e. scalar weigh map) to control the contribution at *ij-th* location in *k-th* image.

*B. Pyramid Generation and Fusion*

Researchers have used multi-resolution approaches to fuse and modify the details at multiple scales called multi-resolution decomposition [23,47] that overcome the problem of seam and contrast reversal artifact produced during image fusion process. In present approach, input image is decomposed in to various spatial bands called Laplacian pyramid and manipulated based on local entropy measure. Only the regions having maximum details are preserved in the final fused image.

Gaussian pyramid of local entropy computed across input images acts as scalar weighting function that controls the contribution of pixels from the multiple exposures. The Laplacian pyramid of image $LI_k^l$ multiplied with the corresponding Gaussian pyramid of scalar weights $GW_k^l$ and summing over $k$ yield modified Laplacian pyramid $L^{*l}$:

$$LI_k^{*l} = -\sum_{k=1}^{N} LI_{ij,k}^{l} GW_{ij,k}^{l}$$

where $l$ and $k$ denotes the level of Gaussian pyramid and input image. The original image that contains well exposed pixels is reconstructed from $L^{*l}$ by expanding each level and summing.

3.  **RESULTS AND DISCUSSION**

The proposed algorithm is implemented on MATLAB-2013a. The results generated from proposed approach yields natural contrast in fused image. The proposed algorithm is tested on a different image data sets captured at different exposure level. It is demonstrated that the results are better than recently proposed tone mapping and multi-exposure image fusion approaches.

Figures 2b, and 3b illustrates the proposed results for "Store" and "Igloo" image data sets, respectively, which illustrate that our results preserve details in brightly and poorly illuminated areas with better color details. To generate results 3x3 block size is fixed for the computation of weight function governed by local entropy metric. The auto-results are illustrated in Figures 2c, 3c, and the exposure fusion results produced by [56] are shown in Figures 2d, 3d. The results proposed by [56] are able to produce better details but yields less color details. It is observed from Figure 3b that the complete light variations are preserved accurately without producing artifacts. The details present in highlights and shadows of "Store" image data sets results (see Figure 2b) are perfectly preserved in proposed fused image.

The results produced by Mertens's et al. [23] are shown in Figures 4, and 5. It is observed that proposed results are able to choose best part of multi-exposure input data sets and the fused image is realistic-looking image that is much nearer to what human eye originally saw. Therefore both details under highlights and shadows of "House" and "Venice" (Figure 4a and 5a) are produced simultaneously (Fused images are shown in Figure 4b and 5b). We can observe that optimal contrast is produced in the fused image. Figure 5b illustrates that proposed results yield more details in the under-exposed region of boat and shadow areas of buildings, which are not present in the results generated by [23].

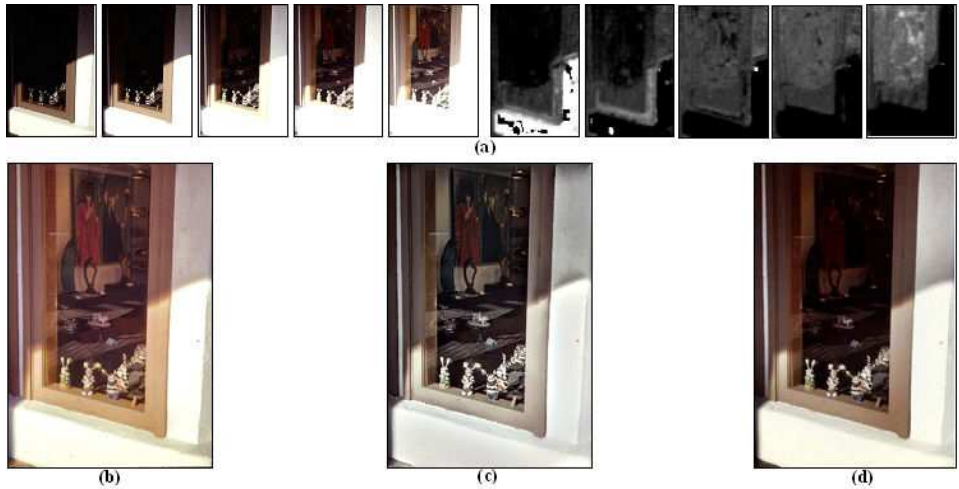

**Fig. 2.** Comparison of "Store" image data sets. (a) Input multi-exposure images (left six) and normalized entropy weight function (right six); (b) Proposed results; (c) auto-result; (d) results by [56]. Input images courtesy of Shree Nayar.

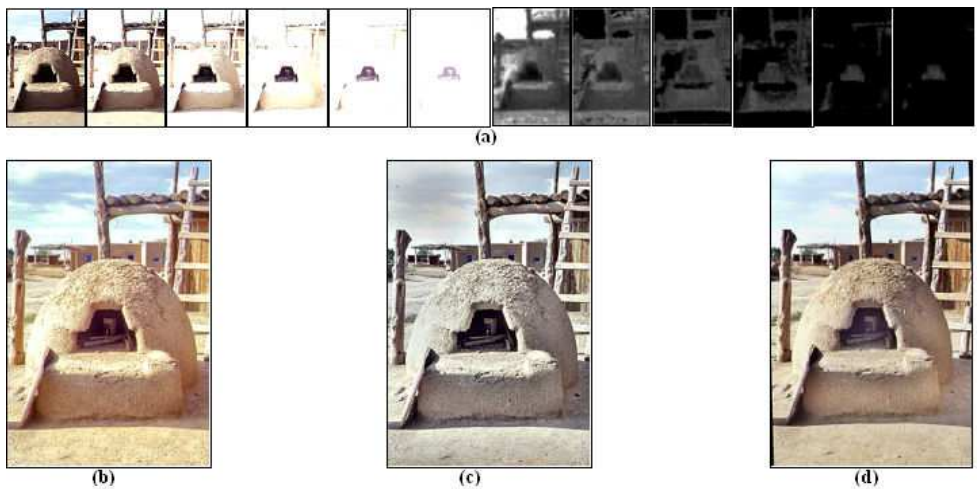

**Fig. 3** Comparison of "Igloo" image data sets. (a) Input multi-exposure images (left six) and normalized entropy weight function (right six); (b) Proposed results; (c) auto-result; (d) results by [56]. Input images courtesy of Shree Nayar.

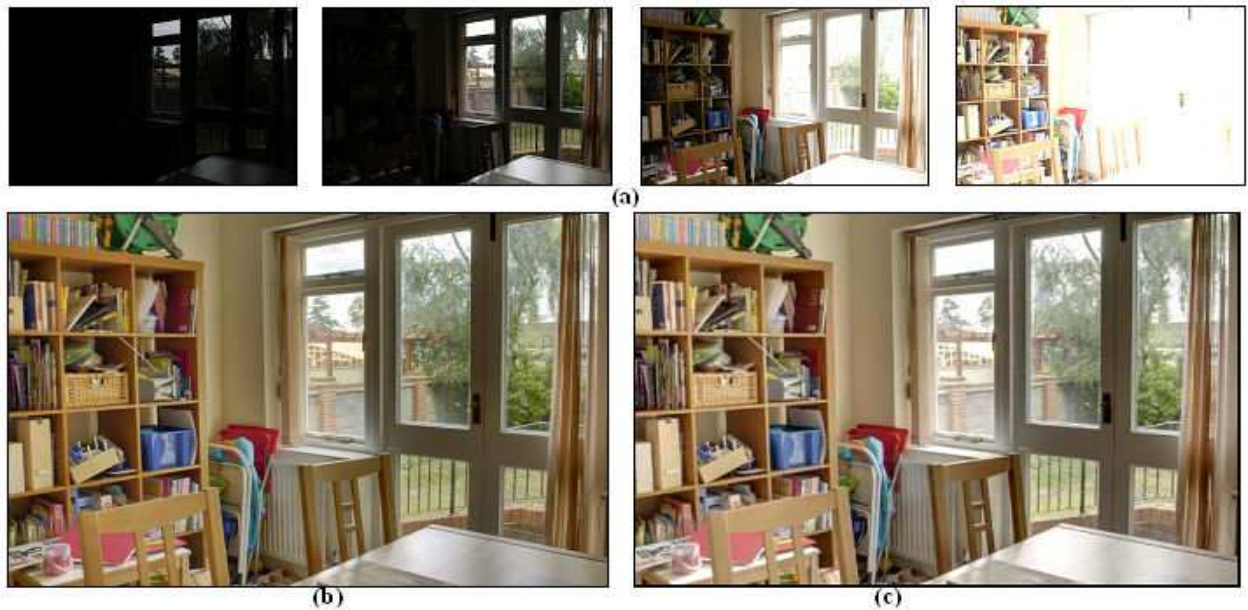

**Fig. 4.** Comparison of "House" image data sets. (a) Input multi-exposure images; (b) Proposed results; (c) results by [23]. Input images courtesy of Jacques Joffre.

To check the effectiveness of present method the proposed results are compared with tonemapped images. Figure 6b illustrates the proposed results of the "Belgium House" image data set. The tone-mapped results of HDR image are illustrated in Figure 6c and 6d. The perceptually driven works proposed by [57] and low curvature image simplifier (LCIS) hierarchical decomposition [58] are taken for comparison. The result proposed by [57] and LCIS [58] are taken original research article for comparison. It can be observed from Figure 6b that proposed fused image is able to preserve finer details and yields natural contrast. The human visual system adaptation property is utilized for dynamic range compression and the tone-mapped image was found to suffer from halo artifacts. Moreover, the output image does not produce good chrominance details (see Figure 6c).

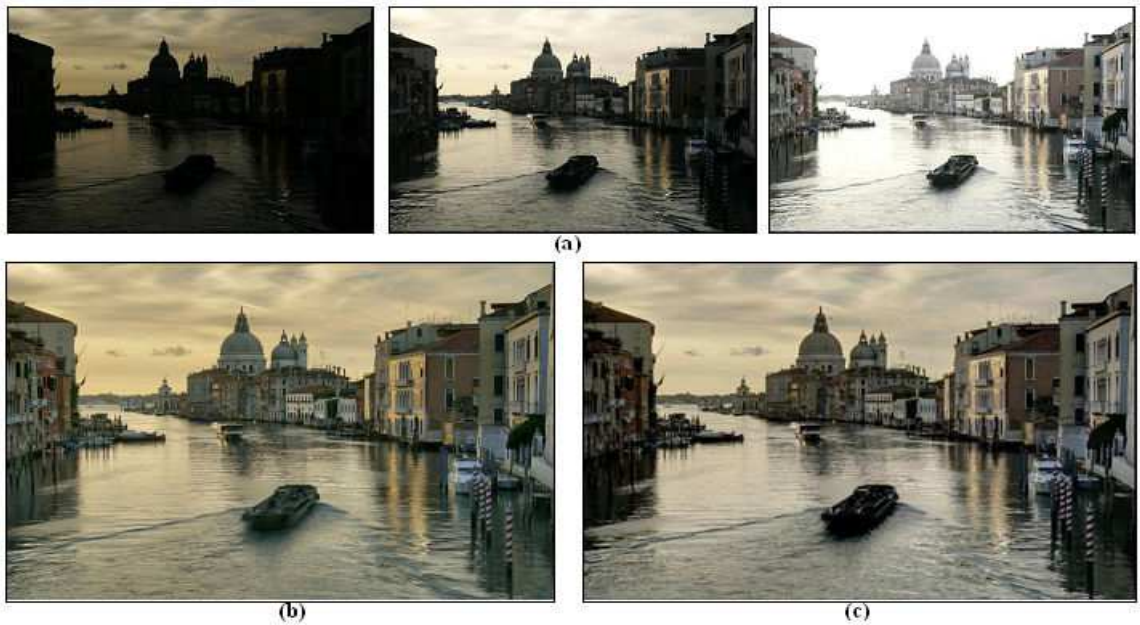

**Fig. 5.** Comparison of "Venice" image data sets. (a) Input multi-exposure images; (b) Proposed results; (c) results by [23]. Input images courtesy of Jacques Joffre.

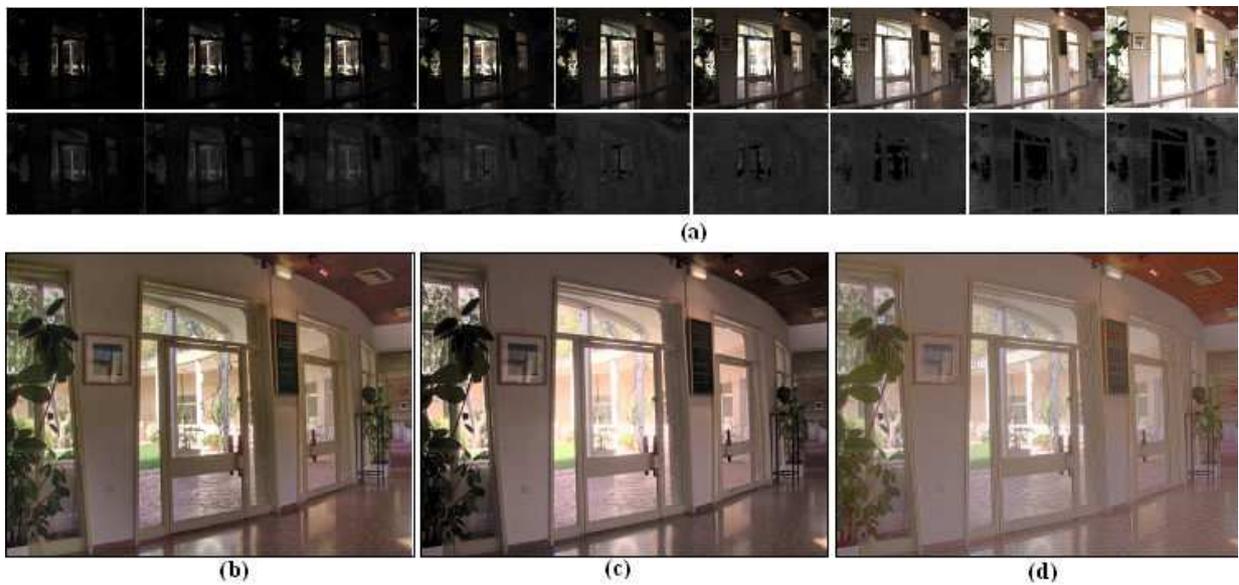

**Fig. 6.** Comparison of "Belgium House" results with popular tone mapping techniques. (a) Input images & corresponding local entropy weight maps (b) Proposed results; (c) Results of Ward Larson et al. [11] and (d) Results of LCIS method [38]. (Input sequence courtesy of Dani Lischinski.)

## 4. SUMMARY

Multi-exposure image fusion is presented to preserve under-exposed and overexposed regions present in the real world scene. In present approach local entropy measure is utilized for computing weight map function. Based on these weight map function the modified Laplacian pyramid is competed to generate fused image across input images. It is presented that the scalar weight maps computed using local entropy is useful to choose well-exposed region across input images. Moreover, it is presented that CLAHE as preprocessing operator is suitable to handle inconsistencies across input multi-exposure images used in fusion process.